\documentclass[letterpaper]{article}

\usepackage{aaai2027}

\nocopyright

\usepackage[hyphens]{url}
\usepackage{graphicx}
\usepackage{natbib}
\usepackage{caption}
\usepackage{amsmath}
\usepackage{amssymb}
\usepackage{booktabs}
\usepackage{tabularx}
\usepackage{array}
\usepackage{multirow}
\usepackage{pifont}
\usepackage{colortbl}
\usepackage{float}
\usepackage{dblfloatfix}
\usepackage{algorithm}
\usepackage{algpseudocode}

\newcommand{\cmark}{\ding{51}}

\title{
ProtoRAG: Prototype-Based Retrieval Augmentation for Few-Shot
Fine-Grained Remote Sensing Object Detection
}

\author{
Jian Wang,\textsuperscript{\rm 1}
Yuxiang Hong,\textsuperscript{\rm 2}
Chufeng Zhou,\textsuperscript{\rm 1}
Chao Pang,\textsuperscript{\rm 1}
Xiaokang Zhang\textsuperscript{\rm 1}\corresponding
}

\affiliations{
\textsuperscript{\rm 1}Wuhan University, Wuhan, China\\
\textsuperscript{\rm 2}East China Normal University, Shanghai, China
}

\begin{document}

\maketitle

\begin{abstract}
Few-shot fine-grained object detection (FGOD) in remote sensing imagery is challenging because limited annotations must support both object localization and discrimination among visually similar subcategories. Although multimodal large language models (MLLMs) provide strong coarse object localization, they lack explicit visual evidence for reliable fine-grained recognition. To address this limitation, we propose ProtoRAG, a prototype-based retrieval-augmented framework that decouples coarse localization from fine-grained recognition by equipping MLLMs with an external object-level visual memory. To construct a reliable visual memory from limited support samples, we introduce Discriminative Prototype Space Learning (DPSL), which encourages discriminative and prototype-stable representations through supervised contrastive learning and prototype-consistency regularization. We further develop an uncertainty-guided candidate-constrained reasoning strategy that augments MLLMs with retrieved candidate-specific visual references and invokes multimodal reasoning only for ambiguous instances. Extensive experiments show that ProtoRAG consistently surpasses representative baselines in nine few-shot settings, outperforming the strongest baselines by 14.80, 2.27, and 4.04 mAP$_{50}$ on MAR20, HRSC2016, and FAIR1M-2.0, respectively.

\end{abstract}
\section{Introduction}
Fine-grained object detection (FGOD) aims to localize objects while distinguishing their subordinate categories, such as different aircraft variants or ship types. Unlike conventional object detection, fine-grained categories often share highly similar global appearances, with discriminative cues residing in subtle local structures, geometric configurations, and component layouts. Capturing these cues is particularly challenging in remote sensing imagery, where objects are typically small and exhibit large variations in orientation, scale, imaging conditions, and background clutter. As a result, remote sensing FGOD relies on extracting highly discriminative visual evidence from inherently limited object-level observations~\cite{chu2024fine,sun2022fair1m}.

Existing remote sensing FGOD methods mainly improve category discrimination through part-aware modeling, global-local interaction, multi-scale reasoning, and background suppression~\cite{ouyang2022multigranularity,guan2023aircraft,wang2024relevance,cheng2023sfrnet,ouyang2023pcldet}. However, these methods are predominantly developed under full supervision and rely on large numbers of precisely annotated instances. Acquiring such annotations is labor-intensive, requires substantial domain expertise, and is particularly difficult for rare fine-grained categories. Under sparse supervision, conventional detectors easily overfit, while existing few-shot detectors primarily transfer localization knowledge from base to novel classes rather than learning representations that explicitly capture the subtle visual differences between subordinate categories~\cite{wang2024fpd,liu2024saefsdet,zhou2025pidivit}.

Recent multimodal large language models (MLLMs) have demonstrated remarkable capabilities in open-vocabulary perception, object localization, and visual reasoning~\cite{kuckreja2024geochat,bai2025qwen3}, making them a promising foundation for reducing the annotation burden of remote sensing FGOD. However, coarse semantic understanding does not directly translate into reliable fine-grained recognition. As illustrated in Fig.~\ref{fig:vlm_motivation}, an MLLM can accurately localize an aircraft while still confusing visually similar variants because no explicit visual references are available during inference. This limitation arises because coarse object recognition mainly relies on semantic knowledge acquired through large-scale vision-language pretraining, whereas fine-grained recognition fundamentally depends on comparing subtle object-level structures against representative visual exemplars. In remote sensing imagery, this challenge is further amplified by small object sizes, cluttered backgrounds, and multiple-object interference, making implicit semantic memories insufficient for reliable subordinate-category discrimination.

Since fine-grained recognition depends on object-level visual evidence, a small set of annotated support instances naturally provides explicit references for distinguishing visually similar categories. Organizing these instances as an external visual memory enables retrieving representative exemplars during inference. However, naively storing support features is insufficient. Fine-grained categories often occupy highly overlapping regions in generic feature spaces, while prototypes estimated from only a few support samples are easily affected by support selection and imaging variations. Consequently, effective retrieval requires a visual memory that is both highly discriminative across categories and robust to support diversity. These observations suggest that augmenting MLLMs with explicit visual memory, rather than relying solely on implicit pretrained knowledge, provides a natural solution for few-shot fine-grained recognition.

Motivated by these observations, we propose \textbf{ProtoRAG}, a prototype-based retrieval-augmented approach for few-shot fine-grained object detection in remote sensing imagery. ProtoRAG decouples coarse localization from fine-grained recognition by retrieving candidate-specific visual references from an external prototype memory after object localization. It further learns a discriminative and robust prototype memory from limited support examples to enable reliable retrieval under sparse supervision. During inference, retrieved visual references directly support confident predictions, while ambiguous cases are resolved through MLLM-based visual comparison. Our main contributions are as follows:

\begin{figure}[t]
    \centering
    \includegraphics[width=\linewidth]{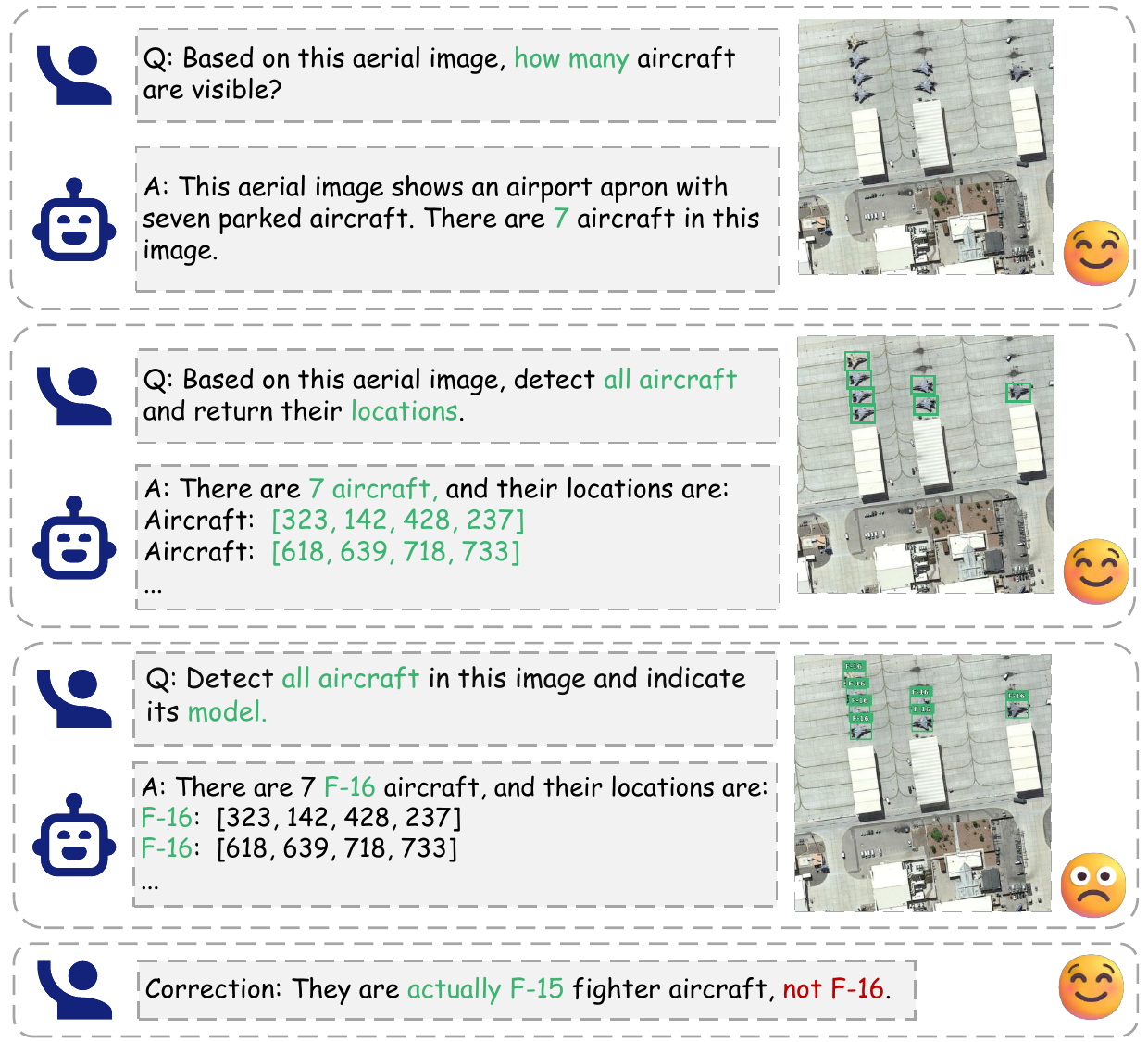}
    \caption{An MLLM can perform coarse object localization but lacks
    task-aligned comparative evidence for distinguishing visually similar
    subordinate categories.}
    \label{fig:vlm_motivation}
\end{figure}


\begin{itemize}

\item We present \textbf{the first retrieval-augmented framework for few-shot remote sensing FGOD}, enabling MLLMs to perform visually grounded fine-grained recognition through an external object-level visual memory.

\item We introduce \textbf{Discriminative Prototype Space Learning (DPSL)}, which learns category-discriminative and prototype-stable visual representations from limited support examples.

\item We develop an uncertainty-guided candidate-constrained reasoning strategy that augments MLLMs with retrieved visual references, invoking multimodal reasoning only for ambiguous predictions.

\item Extensive experiments on MAR20, HRSC2016, and FAIR1M-2.0 demonstrate consistent improvements over conventional, few-shot, open-vocabulary, MLLM-based, and retrieval-based methods.

\end{itemize}


\section{Related Work}
\subsection{Few-Shot Fine-Grained Object Detection}

Fine-grained object detection in remote sensing imagery requires object localization while distinguishing similar subordinate categories. Existing methods enhance fine-grained representations through part-aware modeling, global--local interaction, multi-scale feature learning, and prototypical contrastive learning~\cite{jia2021adaptive,ouyang2022multigranularity,guan2023aircraft,wang2024relevance,ouyang2023pcldet}. Few-shot approaches reduce annotation requirements through prototype learning and task-specific feature adaptation~\cite{wang2024fpd,liu2024saefsdet,zhou2025pidivit}. Nevertheless, these methods follow a detector-centric paradigm in which localization and fine-grained recognition are jointly learned from the same limited annotations. Consequently, sparse supervision must support both spatial localization and subtle inter-category discrimination, limiting the learning of reliable fine-grained representations.

\subsection{MLLM-Based Object Detection}

MLLMs exhibit visual-semantic understanding and can localize objects from coarse-grained category prompts~\cite{kuckreja2024geochat,bai2025qwen3}. Open-vocabulary and image-prompted detectors extend detection to novel categories through textual or visual prompts~\cite{cheng2024yoloworld,pan2025laedino,wu2025vistex}. Despite their prompting mechanisms, these models rely on broadly aligned visual-semantic representations that do not reliably capture the subtle structural differences between fine-grained categories. Retrieval-augmented methods introduce support-derived visual evidence by comparing queries with category prototypes or cached features~\cite{snell2017prototypical,zhang2022tipadapter,liu2026rar}. However, they primarily address image-level recognition or perform retrieval in generic feature spaces, limiting their applicability to few-shot fine-grained object detection.

\begin{figure*}[t]
    \centering
    \includegraphics[width=0.96\linewidth]{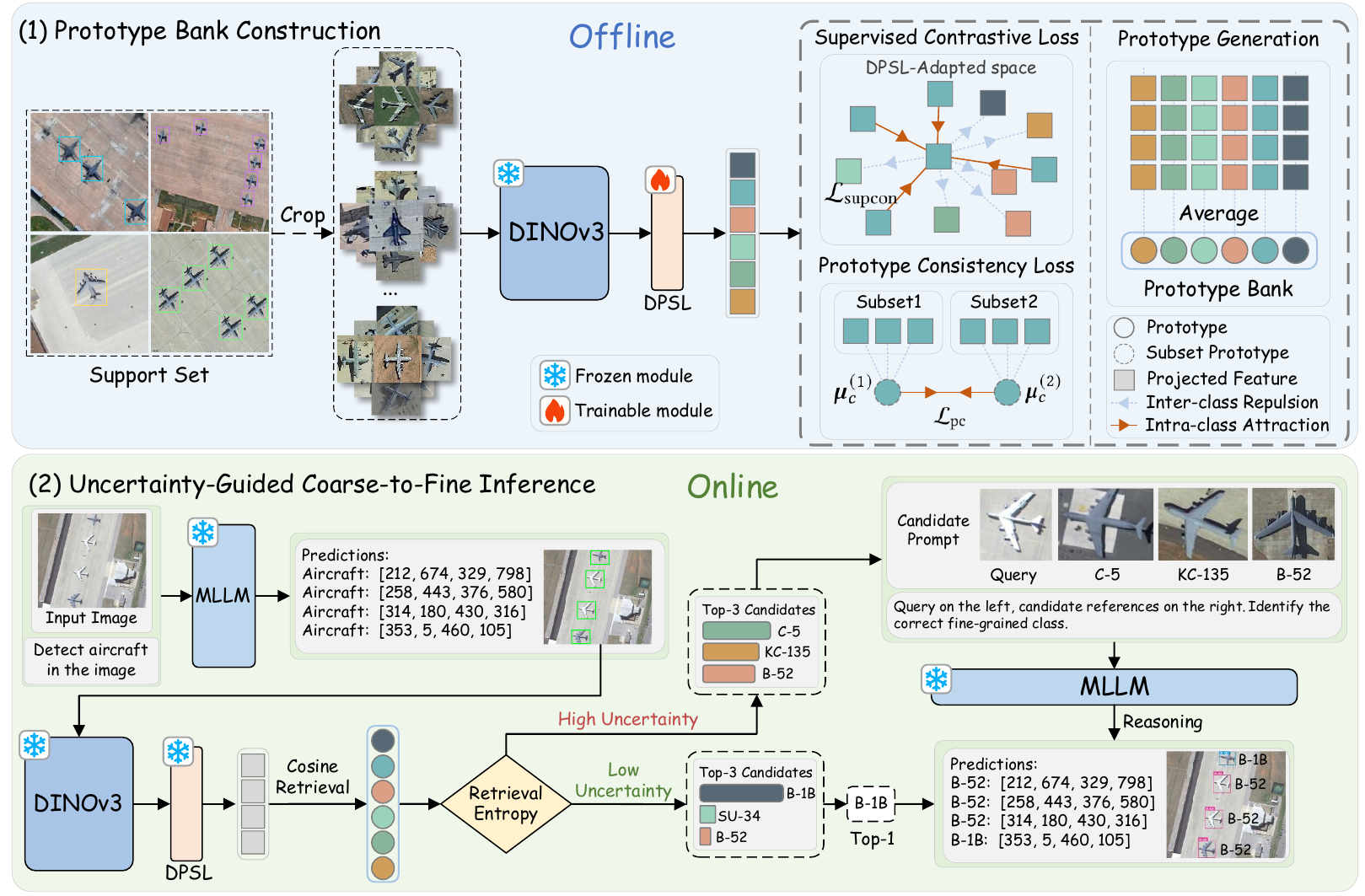}
    \caption{Overview of ProtoRAG. DPSL constructs an object-level prototype
    memory from the support set. An MLLM localizes query targets, which retrieve
    candidate categories and support references. Low-entropy queries use
    direct prototype prediction, whereas high-entropy queries undergo
    candidate-constrained visual comparison.}
    \label{fig:framework}
\end{figure*}

\section{Method}

As shown in Fig.~\ref{fig:framework}, ProtoRAG decouples MLLM-based coarse object localization from few-shot fine-grained recognition through two stages: (1) Discriminative Prototype Space Learning (DPSL) and visual memory construction, and (2) Uncertainty-guided coarse-to-fine inference. The second stage integrates MLLM-based coarse-grained localization, object-level prototype retrieval, and candidate-constrained reasoning into a unified inference pipeline. Only the lightweight feature mapping in DPSL is optimized during training, while the visual encoder and MLLM remain frozen, retaining their pretrained representations while enabling efficient adaptation under limited supervision.

\subsection{Discriminative Prototype Space Learning}
Generic visual representations prioritize broad semantic modeling over the
subtle structural differences required for fine-grained discrimination in
remote sensing imagery. Consequently, visually similar subcategories remain
entangled in the feature space, hindering the construction of category
prototypes that are both discriminative and stable. To address this issue, we
introduce Discriminative Prototype Space Learning (DPSL), which adapts frozen
DINOv3 features to construct reliable fine-grained prototypes.

Given a support object crop $x_i$, the frozen DINOv3 encoder~\cite{simeoni2025dinov3} is utilized to extract its feature
$\mathbf{u}_i=E_{\mathrm{DINOv3}}(x_i)\in\mathbb{R}^{d}$. DPSL produces the
task-adapted embedding:
\begin{equation}
\mathbf{z}_i=\mathrm{DPSL}_{\theta}(\mathbf{u}_i)
=\mathbf{u}_i+\alpha g_{\theta}(\mathbf{u}_i)\in\mathbb{R}^{d},
\end{equation}
where $g_{\theta}:\mathbb{R}^{d}\rightarrow\mathbb{R}^{d}$ is a learnable
residual mapping and $\alpha$ controls the magnitude of the task-specific
correction. The DINOv3 feature, DPSL embedding, and category prototype share
the same dimension $d$. 
Then, DPSL learns the adapted embedding space through two complementary objectives:
instance-level discrimination separates visually similar categories, while
category-level prototype consistency reduces prototype variation caused by
support sampling.

\paragraph{Instance Discrimination}

Consider a mini-batch
\(\mathcal{B}=\{(x_i,y_i)\}_{i=1}^{B}\), where \(y_i\in\mathcal{C}\) is the category label of \(x_i\). For each anchor \(i\), its positive set is
$\mathcal{P}(i)=\left\{p\in\{1,\ldots,B\}\setminus\{i\}\mid y_p=y_i\right\}.$
The supervised contrastive loss is then written as
\begin{equation}
\mathcal{L}_{\mathrm{supcon}}
=
-\frac{1}{|\mathcal{I}|}
\sum_{i\in\mathcal{I}}
\frac{1}{|\mathcal{P}(i)|}
\sum_{p\in\mathcal{P}(i)}
\log
\frac{
\exp(\mathbf{z}_i^{\top}\mathbf{z}_p/\tau)
}{
\sum_{a\neq i}
\exp(\mathbf{z}_i^{\top}\mathbf{z}_a/\tau)
},
\label{eq:supcon}
\end{equation}
where \(\mathcal{I}=\{i\mid |\mathcal{P}(i)|>0\}\) contains anchors with at least one positive sample, and \(\tau\) is the contrastive temperature. This objective reduces intra-category variation while enlarging the margins between visually similar categories.

\paragraph{Category-Level Prototype Consistency}
Although supervised contrastive learning improves instance-level
discrimination, it does not explicitly constrain the variation of category
prototypes constructed from different support subsets. For a specific category
$c\in\mathcal{C}$, let
$\mathcal{S}_c=\{i\mid c_i=c\}$ denote the indices of its support instances,
with $N_c=|\mathcal{S}_c|$. For each category satisfying $N_c\geq 2m$, we
sample two disjoint subsets
$\mathcal{S}_c^{(1)},\mathcal{S}_c^{(2)}\subset\mathcal{S}_c$, each containing
$m$ instances. Their subset prototypes are defined as
$\boldsymbol{\mu}_c^{(t)}
=\frac{1}{m}\sum_{i\in\mathcal{S}_c^{(t)}}\mathbf{z}_i$,
where $t\in\{1,2\}$. Let $\mathcal{C}_m=\{c\in\mathcal{C}\mid N_c\geq 2m\}$ denote the categories
eligible for subset sampling. Prototype consistency is enforced by minimizing
the cosine distance between the two subset prototypes:
\begin{equation}
\mathcal{L}_{\mathrm{pc}}
=
\frac{1}{|\mathcal{C}_m|}
\sum_{c\in\mathcal{C}_m}
\left[
1-
\operatorname{sim}
\left(
\boldsymbol{\mu}_c^{(1)},
\boldsymbol{\mu}_c^{(2)}
\right)
\right].
\label{eq:prototype_consistency}
\end{equation}

The overall DPSL objective is
$\mathcal{L}_{\mathrm{DPSL}}
=\mathcal{L}_{\mathrm{supcon}}+\mathcal{L}_{\mathrm{pc}}$.
The two objectives jointly enhance category discrimination and prototype
stability, producing a task-adapted embedding space for subsequent prototype
construction and retrieval.

\paragraph{Prototype Bank Construction}
After DPSL training, all adapted support embeddings are used to construct the
prototype bank. For each category $c\in\mathcal{C}$, its prototype is computed
as
$\mathbf{p}_c=\frac{1}{N_c}\sum_{i\in\mathcal{S}_c}\mathbf{z}_i
\in\mathbb{R}^{d}$.
Since DPSL explicitly improves category separation and prototype consistency,
the resulting mean embedding provides a compact and stable representation of
the corresponding category. Collecting all category prototypes yields
$\mathbf{P}
=[\mathbf{p}_1,\mathbf{p}_2,\ldots,\mathbf{p}_{|\mathcal{C}|}]^\top
\in\mathbb{R}^{|\mathcal{C}|\times d}$.
During inference, query embeddings are compared with the rows of $\mathbf{P}$
using cosine similarity to retrieve the Top-$K$ candidate categories. We also
cache one support reference crop $x_{\mathrm{ref}}^c$ for each category to
provide visual evidence when candidate-constrained reasoning is triggered.
Together, $\mathbf{P}$ and
$\{x_{\mathrm{ref}}^c\}_{c\in\mathcal{C}}$ constitute the object-level visual
memory, which is constructed exclusively from the support set and fixed before
inference.

\subsection{Uncertainty-Guided Coarse-to-Fine Inference}
\label{subsec:coarse_to_fine_inference}

With the object-level visual memory fixed, ProtoRAG performs inference in a
coarse-to-fine manner. The MLLM first localizes target objects according to a
coarse-grained category prompt. Each localized target is then projected into
the same DPSL embedding space as the support instances and matched against the
prototype bank. Retrieval entropy determines whether the most similar
prototype is accepted directly or candidate-constrained MLLM reasoning is
required.

\paragraph{MLLM-Based Object Localization}
The MLLM is provided with the input image and a textual prompt specifying only
the target super-category, such as aircraft or ships. Without using any
fine-grained category names, it localizes all corresponding objects in the
image. We denote the resulting localized targets by
$\widetilde{\mathcal{X}}
=\{\tilde{x}_i\}_{i=1}^{N}$,
where each $\tilde{x}_i$ denotes the target crop extracted from an
MLLM-localized region, and $N$ is the number of localized targets. Since the subsequent
procedure is independently applied to every target, we describe it below for
an arbitrary $\tilde{x}\in\widetilde{\mathcal{X}}$ and omit the target index
for clarity.

\paragraph{Object-Level Prototype Retrieval}
The localized target $\tilde{x}$ is encoded using the frozen DINOv3 encoder and
DPSL, producing
$\mathbf{z}
=\mathrm{DPSL}_{\theta}(E_{\mathrm{DINOv3}}(\tilde{x}))
\in\mathbb{R}^{d}$.
This is the same DPSL representation used for support instances and prototype
construction.

The cosine similarities between $\mathbf{z}$ and all category prototypes form
the fine-grained class-score vector
$\mathbf{q}=[q_c]
\in\mathbb{R}^{|\mathcal{C}|}$,
where
$q_c=\operatorname{sim}(\mathbf{z},\mathbf{p}_c)$
is the prototype retrieval score for category $c$. We rank the entries of
$\mathbf{q}$ and retain the Top-$K$ categories, denoted by
$c_1,c_2,\ldots,c_{K}$.
Here, $K$ denotes the number of retained candidate categories
and is set to $3$ by default. The Top-1 category $c_1$ provides the direct
prototype prediction, while
$\{c_k\}_{k=1}^{K}$ forms the candidate set for uncertainty
estimation and subsequent reasoning.

\paragraph{Retrieval Uncertainty}
Retrieval entropy measures the ambiguity among the Top-$K$
candidate categories. Their retrieval scores are converted into normalized
candidate probabilities using a temperature-scaled softmax:
\begin{equation}
\hat{q}_k=
\frac{\exp(q_{c_k}/\tau_e)}
{\sum_{j=1}^{K}\exp(q_{c_j}/\tau_e)},\\
H=
-\sum_{k=1}^{K}
\hat{q}_k\log\hat{q}_k,
\label{eq:retrieval_entropy}
\end{equation}
where $\tau_e$ is the entropy temperature. Low entropy indicates that one
category dominates the retrieval distribution, whereas high entropy indicates
ambiguity among multiple fine-grained candidates.

\paragraph{Candidate-Constrained MLLM Reasoning}
When $H<\delta$, the prototype retrieval is considered sufficiently confident,
and ProtoRAG directly assigns the Top-1 category $c_1$. When
$H\geq\delta$, the localized target is treated as uncertain and triggers
candidate-constrained reasoning. The MLLM receives the localized target, its
scene context, and one cached support reference for each of the
Top-$K$ candidate categories. 
This mechanism retains efficient prototype prediction for confident targets
and invokes visually grounded MLLM reasoning only when the retrieved
fine-grained categories remain ambiguous.

\paragraph{Fine-Grained Detection Output}
The prototype score corresponding to the final category,
$q_{\hat{c}}=\operatorname{sim}(\mathbf{z},\mathbf{p}_{\hat{c}})$,
is used as its detection ranking score. Applying the same procedure to every
localized target in $\widetilde{\mathcal{X}}$ produces the final fine-grained
detections, each consisting of the target location, predicted category, and
corresponding prototype score. These scores are used for class-aware AP
evaluation.

\begin{table*}[!t]
\centering
\small
\setlength{\tabcolsep}{1mm}

\begin{tabularx}{\textwidth}{
@{}
>{\raggedright\arraybackslash}X|
>{\centering\arraybackslash}p{1.15cm}|
*{4}{>{\centering\arraybackslash}p{0.73cm}}|
*{4}{>{\centering\arraybackslash}p{0.73cm}}|
*{4}{>{\centering\arraybackslash}p{0.73cm}}
@{}
}
\toprule
\multirow{2}{*}{\textbf{Method}}
& \multirow{2}{*}{\textbf{Venue}}
& \multicolumn{4}{c|}{\textbf{MAR20}}
& \multicolumn{4}{c|}{\textbf{HRSC2016}}
& \multicolumn{4}{c}{\textbf{FAIR1M-2.0}} \\
\cmidrule(lr){3-6}
\cmidrule(lr){7-10}
\cmidrule(lr){11-14}
& & \rule{0pt}{1ex}\textbf{ZS}
& \textbf{3} & \textbf{5} & \textbf{10}
& \textbf{ZS} & \textbf{3} & \textbf{5} & \textbf{10}
& \textbf{ZS} & \textbf{3} & \textbf{5} & \textbf{10} \\
\midrule

\rowcolor{gray!15}
\multicolumn{14}{l}{
\hspace{2pt}\textit{\textbf{Standard Detectors}}
} \\

LEGNet~\cite{lu2025legnet}
& ICCVW
& -- & 8.88 & 14.46 & 20.33
& -- & 2.24 & 7.31 & 9.12
& -- & 1.43 & 6.41 & 11.52 \\

LWGANet~\cite{lu2026lwganet}
& AAAI
& -- & 3.84 & 13.23 & 21.90
& -- & 2.97 & 8.63 & 11.75
& -- & 1.45 & 7.22 & 12.46 \\

Strip R-CNN~\cite{yuan2026striprcnn}
& AAAI
& -- & 10.20 & 19.24 & 41.22
& -- & 4.59 & 10.29 & 16.11
& -- & 3.24 & 8.35 & 14.56 \\

\rowcolor{gray!15}
\multicolumn{14}{l}{
\hspace{2pt}\textit{\textbf{Few-Shot Detectors}}
} \\

FPD~\cite{wang2024fpd}
& AAAI
& -- & 13.89 & 13.89 & 30.58
& -- & 10.42 & 13.16 & 24.10
& -- & 7.82 & 9.95 & 12.72 \\

SAE-FSDet~\cite{liu2024saefsdet}
& TGRS
& -- & 12.30 & 18.46 & 40.14
& -- & 10.12 & 11.88 & 27.36
& -- & 7.74 & 8.96 & 14.42 \\

PiDiViT~\cite{zhou2025pidivit}
& ICCV
& -- & 14.71 & 12.97 & 23.64
& -- & 6.45 & 13.76 & 25.14
& -- & 6.34 & 9.13 & 12.83 \\

\rowcolor{gray!15}
\multicolumn{14}{l}{
\hspace{2pt}\textit{\textbf{Open-Vocabulary Detectors}}
} \\

YOLO-World~\cite{cheng2024yoloworld}
& CVPR
& 3.27 & 15.63 & 18.54 & 28.54
& 1.42 & 14.16 & 16.87 & 23.78
& \underline{1.05} & 2.03 & 10.93 & 14.74 \\

LAE-DINO~\cite{pan2025laedino}
& AAAI
& \textbf{3.76} & 17.63 & 21.42 & 40.78
& \textbf{3.24} & 20.35 & 26.25 & \underline{47.37}
& \textbf{1.19} & 2.31 & 11.27 & 17.43 \\

VisTex-OVLM~\cite{wu2025vistex}
& ICCV
& \underline{3.46} & 3.68 & 3.68 & 3.67
& \underline{3.03} & 3.27 & 3.26 & 3.27
& 1.01 & 2.34 & 2.85 & 2.87 \\

\rowcolor{gray!15}
\multicolumn{14}{l}{
\hspace{2pt}\textit{\textbf{Retrieval-Augmented Methods}}
} \\

DINOv3-kNN~\cite{cover1967nearest}\textsuperscript{\ensuremath{\dagger}}
& --
& -- & 51.18 & 59.27 & 67.14
& -- & 24.03 & 29.53 & 32.46
& -- & 12.77 & 15.84 & \underline{18.43} \\

ProtoNet~\cite{snell2017prototypical}\textsuperscript{\ensuremath{\dagger}}
& NeurIPS
& -- & \underline{55.08} & \underline{60.24} & \underline{69.26}
& -- & \underline{25.32} & \underline{30.56} & 36.40
& -- & \underline{13.68} & \underline{16.75} & 17.31 \\

Tip-Adapter~\cite{zhang2022tipadapter}\textsuperscript{\ensuremath{\dagger}}
& ECCV
& -- & 27.07 & 31.23 & 35.25
& -- & 16.18 & 21.37 & 27.83
& -- & 10.87 & 12.66 & 15.44 \\

RAR-Adapted~\cite{liu2026rar}\textsuperscript{\ensuremath{\dagger}}
& TIP
& -- & 39.80 & 42.24 & 46.26
& -- & 19.21 & 26.84 & 29.48
& -- & 12.58 & 14.75 & 16.86 \\

\midrule
\textbf{ProtoRAG}
& --
& -- & \textbf{58.82} & \textbf{73.76} & \textbf{84.06}
& -- & \textbf{25.39} & \textbf{40.92} & \textbf{49.64}
& -- & \textbf{16.59} & \textbf{20.14} & \textbf{22.47} \\

\bottomrule
\end{tabularx}

\caption{Comparison on MAR20, HRSC2016, and FAIR1M-2.0 under zero-shot
and few-shot settings. Results are mAP$_{50}$ (\%). The best and
second-best results are shown in \textbf{bold} and \underline{underlined},
respectively. \textsuperscript{\ensuremath{\dagger}} These baselines are
adapted to the same DINOv3--Qwen3-VL detection pipeline.}
\label{tab:main_results}
\end{table*}



\section{Experiments}
\subsection{Experimental Setup}
\paragraph{Datasets}
We evaluate on MAR20~\cite{wenqi2023mar20},
HRSC2016~\cite{liu2017high}, and FAIR1M-2.0~\cite{sun2022fair1m}.
MAR20 contains 20 aircraft types; its labels (A1--A20) are replaced
by corresponding aircraft names in MLLM prompts. We select 14 ship
categories from HRSC2016 and retain the aircraft and ship subcategories of
FAIR1M-2.0.
\paragraph{Few-Shot Protocol}
We evaluate $K\in\{3,5,10\}$ using shared support splits generated with random seed 42, with each split containing at least $K$ annotated images per category. ProtoRAG builds its visual memory from annotated objects, while detector baselines use the same images and annotations. Zero-shot methods do not use support data. 

\paragraph{Evaluation Metrics} All methods are evaluated using horizontal bounding boxes. End-to-end detection is evaluated by mAP$_{50}$. Retrieval is evaluated by Top-1 and Top-3 accuracy, and coarse localization by precision, recall, and F1 at IoU $=0.5$. ProtoRAG ranks detections using the prototype similarity of the final assigned class, while detector baselines use their native confidence scores.

\paragraph{Implementation Details}
We use a frozen DINOv3 ViT-L/16 encoder and
Qwen3-VL-8B-Instruct. DPSL employs a two-layer residual MLP with a hidden dimension of
512 and an output dimension of $d=512$. The residual coefficient is set to
$\alpha=0.3$. We train DPSL for 40 epochs using Adam with a learning rate of
$3\times10^{-4}$ and a batch size of 128. The contrastive temperature is set to $\tau=0.07$. During inference, we retain
$K=3$ candidate categories and set the entropy temperature to
$\tau_e=0.07$. Candidate-constrained MLLM reasoning is triggered when
$H\geq\delta$, with $\delta=0.60$ fixed across all experiments.

\begin{figure*}[!t]
    \centering
    \includegraphics[width=\linewidth]{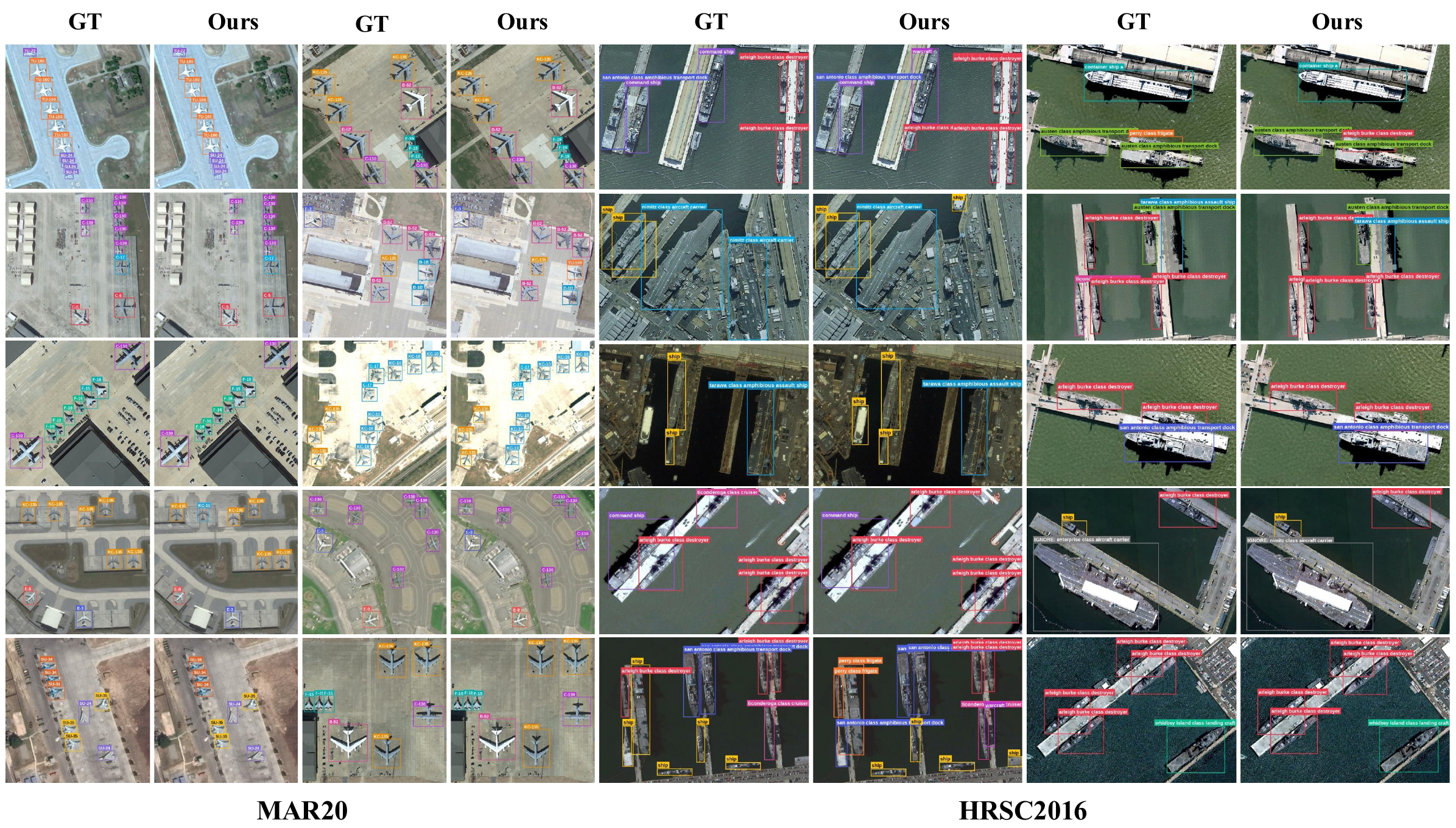}
    \caption{Fine-grained detection results on MAR20 and HRSC2016 under the 10-shot setting.}
    \label{fig:comparison}
\end{figure*}
\subsection{Overall Comparison}
We compare ProtoRAG with representative baselines from four methodological
families. Specifically, LEGNet, LWGANet, and Strip R-CNN are included as
standard detectors; FPD, SAE-FSDet, and PiDiViT as few-shot detectors;
YOLO-World, LAE-DINO, and VisTex-OVLM as open-vocabulary or image-prompted
detectors; and DINOv3-kNN, ProtoNet, Tip-Adapter, and RAR-Adapted as
retrieval-augmented methods. All methods use the same support splits when
applicable. 

Table~\ref{tab:main_results} shows distinct performance trends across different
detection paradigms. Standard and few-shot detectors generally benefit from
additional support images, but remain substantially behind retrieval-based
approaches, particularly on MAR20 and FAIR1M-2.0. This suggests that sparse
annotations provide limited supervision for jointly learning object
localization and subtle fine-grained category boundaries. Meanwhile,
open-vocabulary detectors provide zero-shot detection capability, but their
adaptation to visual support remains uneven. In particular, the nearly
unchanged performance of the image-prompted VisTex-OVLM across different shot
settings indicates that generic image prompts do not readily capture
domain-specific fine-grained distinctions.

Furthermore, the controlled retrieval baselines achieve stronger performance
than most detector-based methods by combining Qwen3-VL localization with
DINOv3 visual matching. For example, ProtoNet reaches 69.26 mAP$_{50}$ on
MAR20 under the 10-shot setting, demonstrating the effectiveness of comparing
localized objects with visual prototypes. However, the relative performance of
DINOv3-kNN, ProtoNet, Tip-Adapter, and RAR-Adapted varies across datasets and
shot settings. This variation indicates that retrieval in a generic
representation space does not consistently produce discriminative and stable
fine-grained prototypes.

Against these baselines, ProtoRAG achieves the best performance across all nine
few-shot evaluations. At $K=10$, it surpasses the strongest competing method
on MAR20, HRSC2016, and FAIR1M-2.0 by 14.80, 2.27, and 4.04 mAP$_{50}$ points,
respectively. More importantly, the controlled retrieval baselines share the
same DINOv3--Qwen3-VL pipeline with ProtoRAG; therefore, the consistent
improvements over this group cannot be attributed solely to MLLM localization.
Instead, they demonstrate the effectiveness of learning discriminative and
stable prototypes and selectively reasoning over ambiguous candidates. The qualitative results in Fig.~\ref{fig:comparison} further show that ProtoRAG assigns accurate fine-grained labels to similar localized targets.

\subsection{Ablation Studies}
\begin{figure}[H]
    \centering
    \includegraphics[width=\linewidth]{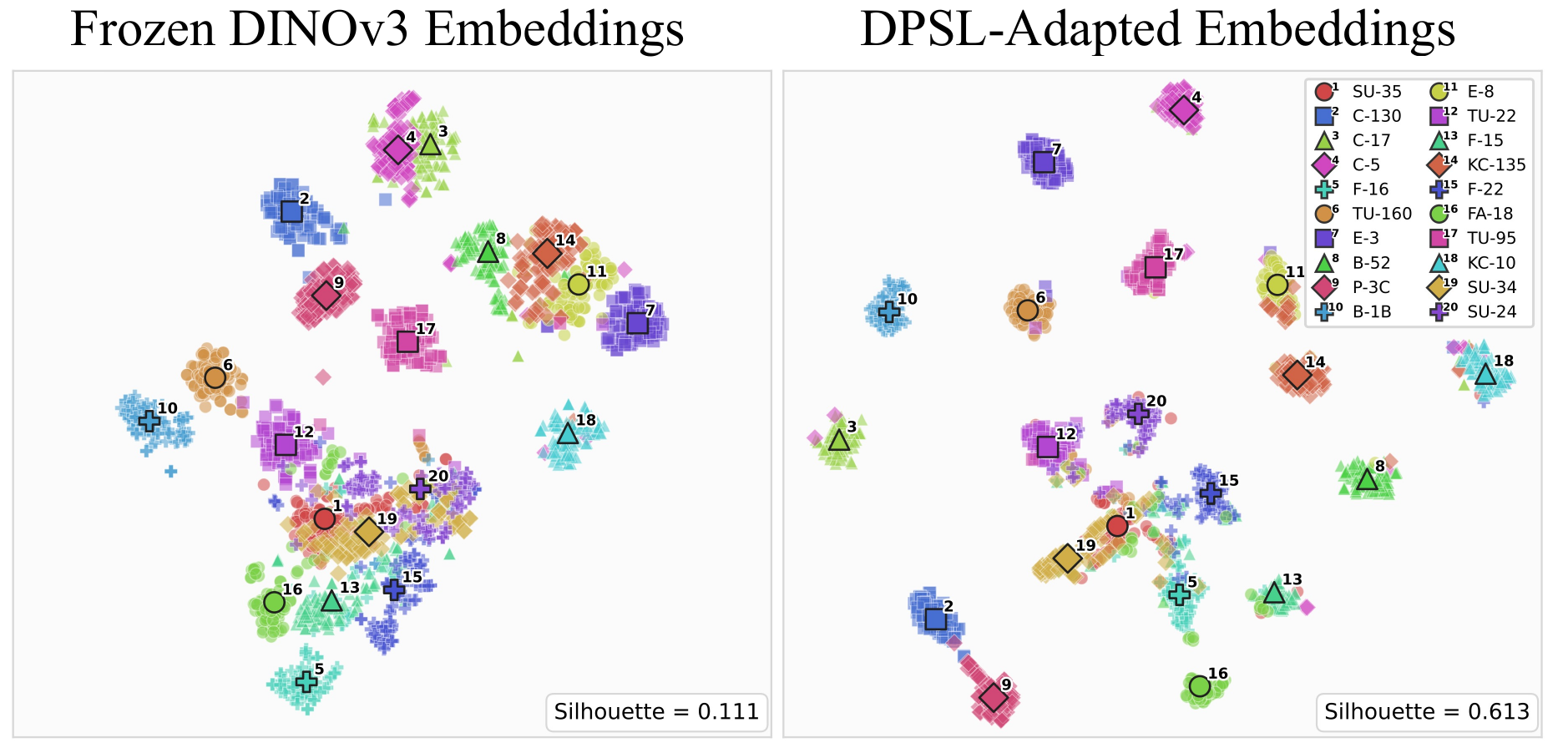}
    \caption{t-SNE visualization of feature discriminability}
    \label{fig:tsne}
\end{figure}

\paragraph{Contribution of Individual Components}
All ablations are conducted on MAR20 under the 10-shot setting. We incrementally evaluate the main components of ProtoRAG: prototype retrieval (PR), which constructs prototypes from frozen DINOv3 support features, Discriminative Prototype Space Learning (DPSL), and uncertainty-guided reasoning (UGR). ZS denotes zero-shot prediction. As shown in Table~\ref{tab:ablation_components}, introducing PR increases
mAP$_{50}$ from 6.34 to 71.08, demonstrating the value of prototype-based
visual matching for fine-grained recognition. DPSL further improves the result
by 12.38 points to 83.46 by learning a more discriminative and stable prototype
space, as visualized in Fig.~\ref{fig:tsne}. Finally, UGR raises mAP$_{50}$ to
84.06 by selectively invoking candidate-constrained MLLM reasoning for
uncertain targets.

\begin{table}[H]
\vspace{0pt}
\centering
\footnotesize
\setlength{\tabcolsep}{5pt}
\renewcommand{\arraystretch}{0.95}
\begin{tabular}{@{}cccc@{\hspace{7pt}}c@{}}
\toprule
\multicolumn{4}{c}{\textbf{Proposed Components}}
& \textbf{Metric} \\
\cmidrule(lr){1-4}
\cmidrule(l){5-5}
\textbf{ZS} & \textbf{PR} & \textbf{DPSL} & \textbf{UGR}
& \textbf{mAP$_{50}$} \\
\midrule
\cmark &        &        &        & 6.34 \\
\cmark & \cmark &        &        & 71.08 \\
\cmark & \cmark & \cmark &        & 83.46 \\
\cmark & \cmark & \cmark & \cmark & \textbf{84.06} \\
\bottomrule
\end{tabular}
\caption{Contribution of the ProtoRAG components.}
\label{tab:ablation_components}
\end{table}

\paragraph{DPSL Training Objectives}
Table~\ref{tab:ablation_loss} evaluates the training objectives of DPSL. $\mathcal{L}_{\mathrm{supcon}}$ achieves 73.05/88.15 Top-1/Top-3 accuracy, while $\mathcal{L}_{\mathrm{pc}}$ obtains 84.92/94.29. Joint optimization further improves these results, demonstrating the complementary effects of instance-level discrimination and category-level prototype stabilization.

\begin{table}[H]
\centering
\begin{tabular}{lcc}
\toprule
\textbf{Training Objective} &
\textbf{Top-1 Acc.} &
\textbf{Top-3 Acc.} \\
\midrule
$\mathcal{L}_{\mathrm{supcon}}$
& 73.05 & 88.15 \\
$\mathcal{L}_{\mathrm{pc}}$
& 84.92 & 94.29 \\
$\mathcal{L}_{\mathrm{supcon}}+\mathcal{L}_{\mathrm{pc}}$
& \textbf{87.24} & \textbf{95.29} \\
\bottomrule
\end{tabular}
\caption{Effect of the DPSL training objectives on retrieval. Retrieval accuracy is reported as Top-1/Top-3 (\%).}
\label{tab:ablation_loss}
\end{table}

\begin{figure*}[!t]
    \centering
    \includegraphics[width=0.94\linewidth]{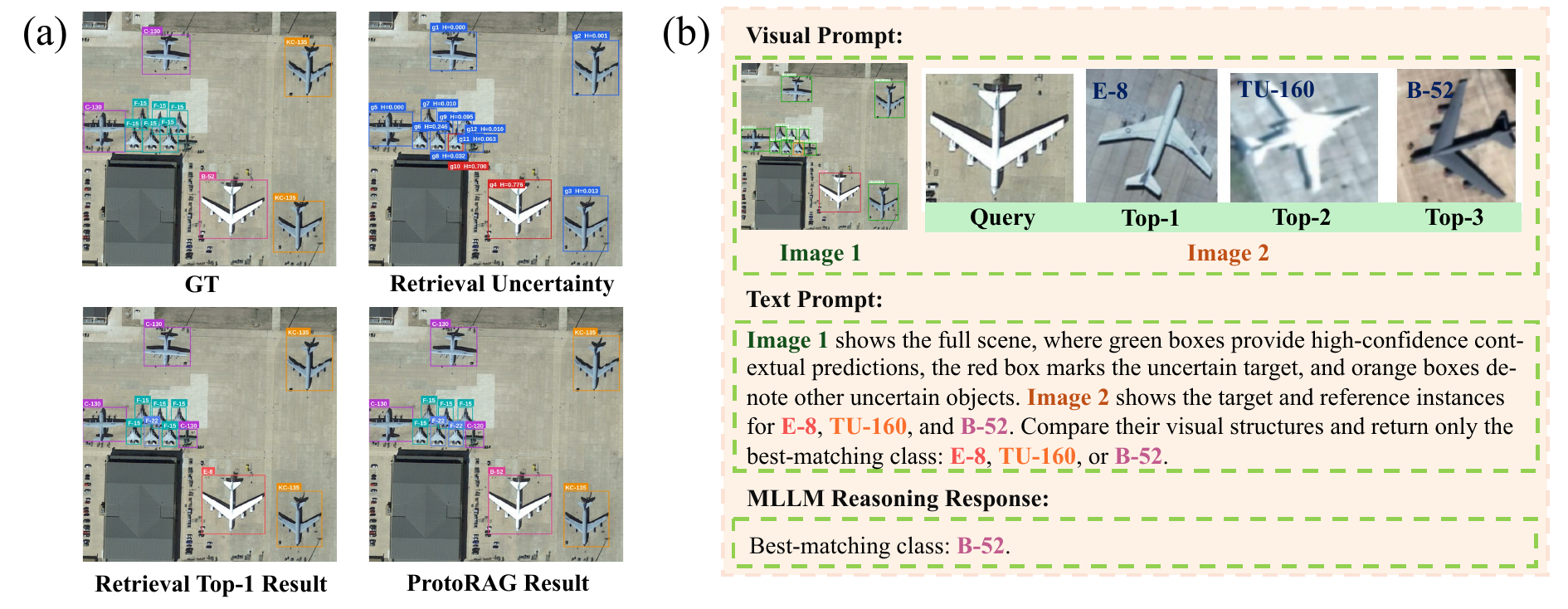}
    \caption{Qualitative illustration of uncertainty-guided reasoning.
(a) Retrieval entropy identifies ambiguous detections.
(b) Candidate-constrained MLLM reasoning corrects an erroneous Top-1
prediction using retrieved visual references.}
    \label{fig:entropy-guided reasoning}
    \vspace{-8pt}
\end{figure*}

\paragraph{Backbone Generalization and DPSL Efficiency}
Table~\ref{tab:r1_backbone} evaluates DPSL with six ViT backbones~\cite{oquab2024dinov2,dosovitskiy2021image} to verify that its effectiveness is not limited to DINOv3 ViT-L/16. Although the large-scale pretrained visual backbones already provide meaningful retrieval representations, DPSL consistently improves their Top-1 and Top-3 accuracy. Moreover, DPSL introduces only 0.43\%--1.23\% trainable parameters and requires minimal training time, demonstrating efficient adaptation across different backbones.

\begin{table}[H]
\centering
\setlength{\tabcolsep}{2.5pt}
\footnotesize
\renewcommand{\arraystretch}{0.95}
\begin{tabular}{lcccc}
\toprule
\textbf{Backbone} &
\begin{tabular}[c]{@{}c@{}}\textbf{w/o DPSL}\\ \textbf{Top-1/Top-3}\end{tabular} &
\begin{tabular}[c]{@{}c@{}}\textbf{w/ DPSL}\\ \textbf{Top-1/Top-3}\end{tabular} &
\begin{tabular}[c]{@{}c@{}}\textbf{Param.}\\ \textbf{Ratio}\end{tabular} &
\begin{tabular}[c]{@{}c@{}}\textbf{Train}\\ \textbf{Time (s)}\end{tabular} \\
\midrule
DINOv2-B & 58.58 / 78.32 & 75.98 / 91.30 & 1.21\% & 34.12 \\
DINOv2-L & 64.97 / 84.77 & 81.27 / 92.79 & 0.43\% & 34.64 \\
DINOv3-B & 64.60 / 82.00 & 77.85 / 91.46 & 1.23\% & 34.34 \\
DINOv3-L & 76.23 / 89.36 & 87.24 / 95.29 & 0.43\% & 34.75 \\
ViT-B\textsuperscript{\ensuremath{\dagger}} & 39.09 / 66.40 & 62.58 / 85.34 & 1.22\% & 34.16 \\
ViT-L\textsuperscript{\ensuremath{\dagger}} & 40.01 / 66.99 & 65.99 / 87.72 & 0.43\% & 37.30 \\
\bottomrule
\end{tabular}
\caption{Backbone generalization and DPSL training cost. 
Param. Ratio is the ratio of trainable DPSL parameters to backbone parameters. \textsuperscript{\ensuremath{\dagger}} denotes  pretrained on ImageNet-1K.}
\label{tab:r1_backbone}
\vspace{-4pt}
\end{table}

\paragraph{Impacts of MLLM Choices}
ProtoRAG leverages the general coarse-localization capability of MLLMs~\cite{wang2025internvl35,bai2025qwen25vl}. As shown in Table~\ref{tab:ablation_vlm}, increasing coarse-detection F1 from 58.59 to 94.94 improves the final mAP$_{50}$ from 46.78 to 84.06, highlighting the importance of reliable localization for downstream fine-grained retrieval.
\begin{table}[H]
\centering

\setlength{\tabcolsep}{2.5pt}
\footnotesize
\renewcommand{\arraystretch}{0.95}
\begin{tabular}{lcccc}
\toprule
\multirow{2}{*}{\textbf{MLLM}}
& \multicolumn{3}{c}{\textbf{Coarse}}
& \textbf{Fine-Grained} \\
\cmidrule(lr){2-4} \cmidrule(lr){5-5}
& \textbf{Precision} & \textbf{Recall} & \textbf{F1}
& \textbf{mAP$_{50}$} \\
\midrule
InternVL3.5-8B  & 58.73 & 58.46 & 58.59 & 46.78 \\
Qwen2.5-VL-7B  & 86.21 & 70.43 & 77.53 & 57.92 \\
Qwen3-VL-8B    & \textbf{93.10} & \textbf{96.85} &
\textbf{94.94} & \textbf{84.06} \\
\bottomrule
\end{tabular}
\caption{Effect of different MLLMs on coarse detection and final
fine-grained detection.}
\label{tab:ablation_vlm}
\vspace{-4pt}
\end{table}

\paragraph{Impacts of Uncertainty-Guided Reasoning}
Table~\ref{tab:entropy_effect_mar20} shows that the reasoning trigger rate
decreases from 14.22\% at $K=3$ to 6.34\% at $K=10$, indicating that larger
support sets produce more confident retrieval results. Nevertheless,
uncertainty-guided reasoning improves mAP$_{50}$ by 0.99, 0.80, and 0.60 points
at $K=3$, 5, and 10, respectively. As illustrated in
Fig.~\ref{fig:entropy-guided reasoning}, high retrieval entropy identifies
difficult instances with ambiguous candidate scores. In this example, direct
Top-1 retrieval misclassifies a B-52 as E-8, whereas candidate-constrained
MLLM reasoning compares the query with the retrieved visual references and
recovers the correct B-52 category. This demonstrates that retrieval entropy
effectively routes uncertain instances to fine-grained reasoning.

\begin{table}[H]
\centering
\footnotesize
\setlength{\tabcolsep}{3.5pt}
\begin{tabular}{cccc}
\toprule
\textbf{$K$} &
\textbf{Trig. (\%)} &
\textbf{Corr. (\%)} &
\textbf{mAP$_{50}$ (Top-1 $\rightarrow$ +UGR)} \\
\midrule
3  & 14.22 & 60.29 &
57.83 $\rightarrow$ 58.82\textsuperscript{+0.99} \\
5  &  9.62 & 66.60 &
72.97 $\rightarrow$ 73.76\textsuperscript{+0.80} \\
10 &  6.34 & 66.77 &
83.46 $\rightarrow$ 84.06\textsuperscript{+0.60} \\
\bottomrule
\end{tabular}
\caption{Uncertainty-guided reasoning at different settings. Trig. is the trigger rate; Corr.
is the probability that a prediction changed by triggered reasoning is
corrected. Superscripts are absolute mAP$_{50}$ gains.}
\label{tab:entropy_effect_mar20}
\end{table}



\section{Conclusion} In this paper, we introduce ProtoRAG, a prototype-based retrieval-augmented framework for few-shot fine-grained object detection in remote sensing imagery. ProtoRAG combines MLLM-based coarse-grained localization with task-specific prototype retrieval, where DPSL learns discriminative and stable prototypes from limited support annotations, and retrieval-entropy-based uncertainty selectively triggers
candidate-constrained reasoning for ambiguous targets. Extensive experiments on three challenging benchmarks demonstrate consistent improvements over representative baselines, validating prototype retrieval as an effective approach for adapting MLLMs to remote sensing FGOD.

\bibliography{reference2}

\end{document}